\documentstyle[11pt,mkstyle,twoside]{article}
\title{Factorization of Dempster-Shafer Belief Functions Based on Data}
\shorttitle{Factorization of DS Beliefs from Data}
\newcommand{\Bem}[1]{}

\Bem{

}

\author{Andrzej Matuszewski, Mieczys{\l}aw A. K{\l}opotek} %

\newcommand{\uu}[1]{\mbox{{$\cal #1$}}}
\newcommand{\ax}[1]{{\bf #1}}
\newcommand{\indep}{\bot}

\newcommand{\Mg}[1]{\{#1\}}

\newcommand{\rwk}{\mbox{$\bigcirc\hspace*{-1.9ex}\mbox{\scriptsize\rm R}$}}
\newcommand{\remove}{\rwk}

\date{Warszawa, November 1995}

\begin{document}

\machetitel







\begin{center}
ABSTRACT 
\end{center}

One important obstacle in applying Dempster-Shafer Theory (DST) is its
relationship to frequencies. In particular,  there exist serious
difficulties in finding factorizations of belief functions from data.
In probability theory factorizations are usually 
related to notion of (conditional) independence and their possibility tested
accordingly. However, in DST conditional belief distributions prove to be
non-proper belief functions (that is ones connected with negative
"frequencies"). This makes statistical testing of potential conditional
independencies practically impossible, as no coherent interpretation could be
found so far for negative belief function values. 
In this paper a novel attempt is made to overcome this difficulty.
In the proposal no 
conditional beliefs are calculated, but instead a new measure F is introduced
within the framework of DST, closely related to conditional independence,
allowing to apply conventional statistical tests for detection of
dependence/independence.

\Bem{
\begin{center}
ROZK{\L}ADANIE NA CZYNNIKI  FUNKCJI PRZEKONANIA
TYPU DEMPSTERA-SHAFERA
NA PODSTAWIE DANYCH
\end{center}

Istotn\c{a} przeszkod\c{a} w stosowaniu teorii Dempstera-Shafera (DST)
jest brak jasnej relacji do cz\c{e}sto\'{s}ci. W szczeg\'{o}lno\'{s}ci stwarza 
to
problemy przy znajdywaniu rozk{\l}adu na czynniki funkcji przekonania z 
danych. 
W teorii prawdopodobie\'{n}stwa rozk{\l}ad na czynniki zwykle jest 
zwi\c{a}zany
z~warunkow\c{a} niezale\.{z}no\'{s}ci\c{a} i mo\.{z}liwo\'{s}\'{c} 
dekompozycji badana jest sto\-sow\-ny\-mi
testami niezale\.{z}no\'{s}ci. Ale w DST warunkowe funkcje przekonania 
nie s\c{a} na og\'{o}{\l} koherentne i st\c{a}d nie posiadaj\c{a} 
interpretacji poz\-wa\-la\-j\c{a}\-cej
na dokonywanie test\'{o}w na zgodno\'{s}\'{c} rozk{\l}ad\'{o}w. 
Artyku{\l} przedstawia nowatorsk\c{a} pr\'{o}b\c{e} rozwi\c{a}zania tego 
problemu poprzez       
zdefiniowanie nowej mia\-ry F w ramach DST, kt\'{o}ra jest wsz\c{e}dzie 
nieujemna,
cz\c{e}sto\'{s}ciowo interpretowalna, a jednocze\'{s}nie \'{s}ci\'{s}le 
zwi\c{a}zana z~wa\-run\-ko\-w\c{a} niezale\.{z}no\'{s}ci\c{a}, co umo\.{z}liwia stosowanie -
po pewnych modyfikacjach - znanych procedur statystycznych  do wykrywania
zale\.{z}no\'{s}ci/nie\-za\-le\.{z}\-no\'{s}\-ci mi\c{e}dzy zmien\-ny\-mi.

\newpage


\thispagestyle{empty}

\quad

\newpage

}

\section{Introduction}

The Dempster-Shafer (DS) Theory (DST) or the Theory of Evidence 
 is  considered  by  many  researchers  as   an 
appropriate tool to 
represent various aspects of human dealing with uncertain knowledge, 
especially for representation of partial ignorance.

However, one particular obstacle in applying DST is its relationship to 
frequencies \cite{Wasserman:92ijar}. Though, in general 
a belief function may be derived from frequencies under some particular
database representation \cite{Klopotek:94f}, there exist serious
difficulties in finding factorizations of belief functions from data.

In probability theory and in classical statistics the  factorizations are
usually
related to notion of (conditional) independence and such  possibility is
tested
accordingly. However, in DST conditional belief distributions prove to be
non-proper belief functions (that is ones connected with negative
"frequencies"). This makes statistical testing of potential conditional
independencies practically impossible, as no coherent interpretation could be
found so far for negative belief function values. 

In this paper a novel attempt is made to overcome mentioned difficulty in
that no
conditional beliefs are calculated, but instead a new measure F is introduced
within the framework of DST, closely related to conditional independence,
allowing to apply conventional statistical tests for detection of
dependence/independence.

The paper is structured as follows: First, basic notions of DST are
introduced. Then the problem with emerging negative beliefs is explained. The
new F-measure is defined. The last section explains the way statistical tests
may be used in connection with this F-measure.

\section{
 Dempster Shafer Theory and the Concept of Conditional Independence
}

 The Valuation Based Systems (VBS) framework,
covering common concepts of probability theory, Dempster Shafer theory of
evidence, to some extent also possibility theory, 
 was introduced in
\cite{Shenoy:94}. In
VBS, a domain knowledge is represented by entities called {\it variables } and
{\it valuations}. Further, two operations called {\it combination } and 
{\it marginalization }  are defined on valuations to perform a local
computational method for computing marginals of the joint
valuation.  The basic components of VBS can be characterized as
follows.

\quad\\
{\bf 
Valuations
}\\

 Let  $\uu{X}=\{X_1,X_2,...X_n\}$ be a finite set of variables and $\Theta_i$
be the domain (called also {\it frame}), i.e. a discrete set of possible
values of i-th variable. If h is a finite non-empty set of 
variables then $\Theta(h)$ denotes the Cartesian product of $\Theta_i$  for
$x_i$ in $h$, i.e. $\Theta(h) = \times \{\Theta_i|X_i \in h\}$. For each
subset s of \uu{X} there is a set $D(s)$
called the domain of a valuation. For instance in the case of
probabilistic systems $D(s)$ equals to $\Theta(s)$, while under the
belief function framework $D(s)$ equals to the power set of  $\Theta(s)$,
i.e. $D(s) = 2^{\Theta(s)}$. Valuations, being primitives in the VBS
framework, can be characterized as mappings $\sigma: D(s) \rightarrow
\uu{R}$  where
 \uu{R} stands for
a set of non-negative reals.
 In
the sequel non-specific valuations will be denoted by lower-case Greek letters,
 $\rho$, $\sigma$, $\tau$, and so on.  The set of all valuations will be
denoted by \uu{V}, wheras $\uu{V}_s$ denotes the set of all valuations defined
for the set of variables $s$.

Within the Dempster-Shafer theory of evidence,
valuation  is either the mass function m,
 belief function Bel, plausibility function Pl or commonality function Q
interchangeably. These functions can be uniquely computed one  from
another using
the formulas:
\begin{eqnarray}
& m(\emptyset)=0  \nonumber\\
& Bel(A)=\sum_{B\subseteq A} m(B) \nonumber\\
& Pl (A)=\sum_{B\cap A\neq\emptyset} m(B) \\
& Q  (A)=\sum_{A\subseteq B} m(B) \nonumber
\end{eqnarray}

 Following Shenoy 
\cite{Shenoy:94} we distinguish
three categories of valuations:

\begin{itemize}
\item 
 {\it Proper valuations}, \uu{P}, represent knowledge that is partially
coherent. (Coherent knowledge means knowledge that has well
defined semantics.) This notions plays an important role in the
theory of belief functions: by proper valuation it is understood
a valuation in which everywhere $m(A)\ge 0$.      
\item 
 {\it Normal valuations}, \uu{N}, represent another kind of partially
coherent knowledge. For instance, in Dempster-Shafer theory, a
normal valuation is an m-function whose values sum to 1.
Particularly, the elements of $\uu{P} \cap \uu{N}$ are called  proper normal
valuations; they represent knowledge that is completely coherent
or knowledge that has well-defined semantics.  
We speak about proper mass function, proper belief function, proper
plausibility function and proper commonality function iff 
$\sum_{A\subseteq\Theta(s)}m(A)=1$ and 
$\forall_{A\subseteq\Theta(s)} m(A)\ge 0$.

 \item 
 {\it Positive normal valuations}: it is a subset $\uu{U}_s$ of $\uu{N}_s$
consisting of  all valuations that have unique identities in  $\uu{N}_s$.
For Dempster-Shafer theory this means $m(\Theta(s))>0$.

\end{itemize}

Further there are two types of special valuations:

\begin{itemize}
\item 
 {\it Zero valuations} represent knowledge that is internally
inconsistent, i.e. knowledge whose truth value is always false;
e.g., in Dempster-Shafer theory by zero valuation we understand a
valuation that is identically zero, $m(A)=0$ for every set A. It is assumed
that for each $s \subseteq \uu{X}$ there is at most one valuation $\zeta_s \in
\uu{V}_s$ . The
set of all zero valuations is denoted by~\uu{Z}.
\item 
 {\it Identity valuations}, I, represent total ignorance, i.e. lack
of knowledge. In Dempster-Shafer theory an identity valuation
corresponds so-called vacuous valuation, where  $m(A)=0$ for every set A
except for $m(\Theta(s)=1$. 
\end{itemize}

\quad\\
{\bf Combination
}\\

 By combination we understand a mapping $\otimes :\uu{V} \times \uu{V} 
 \rightarrow \uu{N} \cup  \uu{Z}$ that
satisfies the following six axioms:

\begin{description}
\item[\ax{(C1)}] If $\rho \in \uu{V}_r$ and $\sigma \in \uu{V}_s$ then 
$\rho \otimes  \sigma \in \uu{V}_{r\cup s}$;

\item[\ax{(C2)}] $\rho \otimes  (\sigma \otimes \tau) = 
(\rho \otimes  \sigma) \otimes  \tau$;
 
\item[\ax{(C3)}] $\rho \otimes  \sigma = \sigma \otimes  \rho$; 

\item[\ax{(C4)}] If  $\rho \in \uu{V}_r$ and zero valuation $\zeta_s$ 
exists then $\rho \otimes  \zeta_s \in \uu{V}_{r\cup s}$.

\item[\ax{(C5)}] For each $s \subseteq \uu{X}$ there 
exists an identity valuation $\iota_s \in  \uu{N}_s
\cup \{\zeta_s\}$ such that for each valuation 
$\sigma \in  \uu{N}_s \cup \{\zeta_s\}$, $\sigma \otimes  \iota_s = \sigma$.

\item[\ax{(C6)}] It is assumed that the set $\uu{N}_\emptyset$  consists of
exactly one element denoted $i_\emptyset$ .
\end{description}

 In practice combination of two valuations is implemented 
as follows. Let (+) be a binary operation on \uu{R}. Then 
$(\sigma \otimes  \rho)(x) =
\sigma(x.s) (+) \rho(x.r)$ where $x$ is an element from $D(s)$ and $x.r$, 
$x.s$
stand for the projection (relying upon dropping unnecessary
variables) of $x$ onto the appropriate domain $D(r)$ or $D(s)$. In
 Dempster-Shafer
theory to the Dempster rule of combination. We say that we combine two mass
functions $m_1$, $m_2$ to obtain $m=m_1\oplus m_2$ iff 
\begin{equation}   
m(A)= 
     \frac{\sum_{B\cap C=A} m_1(B) \cdot m_2(C)}
          {\sum_{B\cap C\ne\emptyset} m_1(B) \cdot m_2(C)}
\end{equation}
Combination of Bel, Pl, Q is the combination of the respective m function.
It is worth mentioning that we can compute  $Q=Q_1\oplus Q_2$ 
as 
   $$Q(A)= 
     \frac{Q_1(A)\cdot Q_2(A)}
          {\sum_{B\cap C\ne\emptyset} m_1(B) \cdot m_2(C)}
$$

 In the field of uncertain reasoning combination corresponds to
aggregation of knowledge: when $\rho$ and $\sigma$ represent our knowledge
about variables in subsets $r$ and $s$ of  
\uu{X} then the valuation $\rho \otimes 
\sigma$ represents the aggregated knowledge about variables in $r  \cup  s$.

 If $\rho \otimes  \sigma$ is a zero valuation, we say that $\rho$ and
$\sigma$  are
inconsistent. On the other hand, if $\rho \otimes  \sigma$ is a normal
valuation, then we say that $\rho$ and $\sigma$ are consistent.
Inconsistency in DST appears if
$$          {\sum_{B\cap C\ne\emptyset} m_1(B) \cdot m_2(C)} = 0$$

\quad\\
{\bf Marginalization}\\

 While combination results in knowledge expansion,
marginalization results in knowledge contraction. Let $s$ be a
non-empty subset of  \uu{X}. It is assumed that for each variable X
in $s$ there is a mapping $\downarrow (s - \{X\}):\uu{V}_s \rightarrow
\uu{V}_s-\{X\}$,
called
marginalization to $s - \{X\}$ or deletion of $X$, that satisfies the
six axioms below:

\begin{description}

\item[\ax{(M1)}] Suppose $\sigma \in \uu{V}_s$ and suppose $X, Y \in  s$. 
 Then\\
 $(\sigma ^{\downarrow (s - \{X\})})^{\downarrow (s
- \{X,Y\})} = (\sigma ^{\downarrow (s - \{Y\})}
)^{\downarrow (s - \{X,Y\})}$ ;

\item[\ax{(M2)}] If zero valuation exists, then 
$\zeta_s ^{\downarrow (s - \{X\})} = \zeta_{s-\{X\}}$;

\item[\ax{(M3)}] 
$\sigma ^{\downarrow (s - {X})} \in  \uu{N}$ if and only if 
$\sigma \in  \uu{N}$ ;

\item[\ax{(M4)}] If $\sigma \in  \uu{U}$ 
then $\sigma ^{\downarrow (s - {X})} \in  \uu{U}$;

\item[\ax{(CM1)}] Suppose $\rho \in \uu{V}_r$ and $\sigma \in \uu{V}_s$. 
Suppose $X \not\in  r$ and $X \in  s$. Then

 $(\rho \otimes  \sigma)^{\downarrow ((r \cup  s) - \{X\})} = 
\rho \otimes
\sigma ^{\downarrow ( s - \{X\})}$ 

\item[\ax{(CM2)}] Suppose $\sigma \in  \uu{N}_s$. Suppose 
$r \subseteq s$ and suppose that $\iota$  is an
identity for $\sigma ^{\downarrow r}$. Then

 $\sigma \otimes  \iota = \sigma$. 

\end{description}

 Axiom \ax{M1}   states that if we delete from s, the domain of a
valuation $\sigma \in \uu{V}_s$, two variables, say $X$ and $Y$, then the
resulting valuation defined over the subset $r = s - \{X,Y\}$ is invariant
with respect to
the order of these variables deletion. Particularly, deleting
all variables from the set s we obtain the valuation whose
domain is the empty set (its existence is guaranteed by axiom
\ax{C6}); by axiom \ax{M3} this element equals to $\iota_\emptyset$  if and
only if $\sigma$ is a normal valuation. 

 Axioms \ax{M2 - M4} state that the marginalization preserves
coherence of knowledge.

In the Dempster-Shafer theory marginalization means summing of masses along
deleted dimensions:
\begin{equation}   
m ^{\downarrow p}(A) = \sum_{B; B ^{\downarrow p}=A} m(B)
\end{equation}   
where marginalization of a set of vectors B onto a subset of variables p means
the set of corresponding vectors projected onto subspace p.

\quad\\
{\bf Removal
}\\

 Removal, called also direct difference, is an "inverse"
operation to the combination. Formally, it can be defined as a
mapping $\remove :\uu{V} \times  (\uu{N} \cup  \uu{Z}) 
\rightarrow \uu{N} \cup  \uu{Z}$, that satisfies the
three axioms: 

\begin{description} 
\item[\ax{(R1)}] If  $\sigma \in \uu{V}_s$ and $\rho \in  \uu{N}_r\cup  
\uu{Z}_r$ then 
$\sigma \remove\rho \in  \uu{N}_{r\cup s}\cup  \uu{Z}_{r\cup s}$.
\item[\ax{(R2)}] For each $\rho \in  \uu{N}_r\cup  \uu{Z}_r$ and for each $r
\subseteq \uu{X}$
there exists an identity $\iota_r$ such that $\rho \remove  \rho = 
\iota_r$ .
\item[\ax{(CR)}] If  $\sigma, \tau \in \uu{V}$ and $\rho \in  \uu{N} \cup  
\uu{Z}$ then 
$(\sigma \otimes  \tau)
\remove \rho =  \sigma \otimes  (\tau \remove  \rho)$.
\end{description}

 Note that we can define the (pseudo)-inverse of a normal
valuation by setting  
$\rho ^{-1} = \iota_\emptyset  \remove  \rho$. 

In the Dempster-Shafer theory, 
the removal $Q=Q_1\ominus Q_2$  is defined (by Shenoy) as
   $$Q(A)=  c \cdot 
     \frac{Q_1(A)}{Q_2(A)}
$$
if $Q_2(A)\ne 0$ and 
$$Q(A)=0$$
otherwise; 
where c is a normalization factor for Q.

He defined conditional independence as follows:
Suppose $\tau \in N_w$, suppose r,s,v are disjoint subsets of w. We say that r
and s are conditionally independent given v with respect to  $\tau$,
written as $r \indep_\tau s|v$ iff there exist 
$\alpha_{r \cup v} \in V_{r \cup v}$  and 
$\alpha_{s \cup v} \in V_{s \cup v}$  such that
$$\gamma(r\cup s\cup v)= 
\alpha_{r \cup v} \oplus
\alpha_{s \cup v} $$

In case of DST conditional independence of sets of variables r and s given v
in belief function Bel 
means that there must exist (not necessarily proper and normal) "belief
functions" $Bel_1$ defined over $r\cup v$ and $Bel_2$ defined over $s\cup v$
such that $$Bel = Bel_1 \oplus Bel_2$$ 

Shenoy introduces also the notion of conditional
valuations and particularly of conditional
 belief functions based on the notion of removal. 

We say that $Bel ^{|p}$ is a DS belief function Bel conditioned on the set of
variables p if 
$$Bel ^{|p}=Bel\ominus Bel ^{\downarrow p}$$.

Furthermore, we can easily derive the conclusion that 
In case of DST conditional independence of sets of variables r and s given v
in belief function Bel 
means that
\begin{equation}   
Bel = (Bel ^{\downarrow (r\cup v)})^{|v} \oplus 
(Bel ^{\downarrow (s\cup v)})^{|v} 
\oplus Bel ^{\downarrow v}
\end{equation}

Shenoy writes, however \cite[pp.225-226]{Shenoy:94}:
"Notice that if $\sigma$ and $\rho$ are commonality functions, it is possible
that  $\sigma\ominus\rho$  may  not  be  a  commonality  function 
because condition
... [of non-negativity of mass function] may not be satisfied by
$\sigma\ominus\rho$ In fact, if $\sigma$ is a commonality function for s, and
$r\subseteq s$, then even $\sigma\ominus{\sigma}^{\downarrow r}$ may fail to
be a commonality function. This fact is the reason why we need the concept of
proper valuation as distinct from non-zero and normal valuations in the
general VBS framework. An implication of this fact is that conditionals may
lack semantic coherence 
in the Dempster-Shafer's theory. This is the primary reason why conditionals
are neither natural nor widely studied in the Dempster-Shafer's
belief-function theory". 

\section{The Fundamental Problem of Testing Conditional Independence in DST}

Dempster-Shafer theory of evidence has been frequently criticized for its
unclear relation to frequencies 
\cite{Wasserman:92ijar} However, even if
we have already agreed on a representational model for daatabase founded
belief functions like that in \cite{Klopotek:94f} then we have still serious
problems with search for conditional independence in a database. \\

First of all, as already stated by Shenoy (cited above), conditional belief
functions are in general not coherent belief functions, hence it is
impossible to formulate for them a counterpart in the world of frequencies. \\

Hence one can be tempted to test if the right hand side and the
left hand side belief distributions in the formula 
$$Bel = (Bel ^{\downarrow (r\cup v)})^{|v} \oplus 
(Bel ^{\downarrow (s\cup v)})^{|v} 
\oplus Bel ^{\downarrow v}$$
agree, e.g. a an appropriate $\chi ^2$-test on agreement of cell frequencies
of empirical Bel  distribution and the "theoretical" "expected" distribution 
$(Bel ^{\downarrow (r\cup v)})^{|v} \oplus 
(Bel ^{\downarrow (s\cup v)}){^|v} 
\oplus Bel ^{\downarrow v}$. But as this "expected" distribution may contain
pseudo-belief functions as components, then the whole distribution 
may also have
negative cells and hence impossible to compare as "expected frequency". \\

One may be tempted to seek heuristically for two (proper normal)  belief
functions $Bel_1$ defined over $r\cup v$ and $Bel_2$ defined over $s\cup v$
such that $$Bel = Bel_1 \oplus Bel_2$$ so that the "expected" distribution 
is ensured to be proper normal by the very coherence of both $Bel_1$ and
$Bel_2$. However, as can be seen from the example below, such $Bel_1, Bel_2$
may not exist at all. 

Let us consider the belief function Bel in variables  X,Y,Z having ranges:
X:\Mg{p,q}, Y:\Mg{r,s,t}, Z:\Mg{a,b,c}. 
The belief distribution $Bel$ in X,Y,Z be:
\begin{center}
\begin{tabular}{|l|r|}
\hline
Set (focal points of $Bel$)   & $m(Set)$\\
\hline
\Mg{(p,r,a), (p,r,b)}                                          & 0.160\\
\Mg{(p,s,a), (p,s,b), (p,t,a), (p,t,b)}                        & 0.040\\
\Mg{(q,r,a), (q,r,b)}                                          & 0.120\\
\Mg{(q,s,a), (q,s,b), (q,t,a), (q,t,b)}                        & 0.030\\
\Mg{(p,r,b), (p,r,c)}                                          & 0.015\\
\Mg{(p,s,b), (p,s,c), (p,t,b), (p,t,c)}                        & 0.060\\
\Mg{(q,r,b), (q,r,c)}                                          & 0.07375\\
\Mg{(q,s,b), (q,s,c), (q,t,b), (q,t,c)}                        & 0.295\\
\Mg{(p,s,b), (p,t,b)}                                          & 0.075\\
\Mg{(q,r,b)}                                                   & 0.13125\\
\hline
TOTAL                                                          & 1.000\\
\hline
\end{tabular} 
\end{center}

It is easily checked that 
$$Bel = (Bel ^{\downarrow \{X,Z\}})^{|Z}  \oplus
(Bel ^{\downarrow \{Y,Z\}})^{|Z} \oplus Bel ^{\downarrow Z}  
$$
Let $Bel_1$ and $Bel_2$ in variables X,Z and Y,Z be two proper belief
functions (that is with non-negative m's) such that $Bel=Bel_1\oplus Bel_2$.
It cannot happen simultaneously that 
$Bel_1$ has any focal point $A$ such that 
$A\downarrow{Z}$ = \Mg{a,b,c} 
and $Bel_2$ has any focal point $B$ such that
$B\downarrow{Z}$ = \Mg{a,b,c} 
because $Bel$ would have to have a   focal point C such that 
$C\downarrow{Z}$ = \Mg{a,b,c}, which is not the case. 

The above fact, 
due to existing focal points,     implies that EITHER
$Bel_1$ must have focal points:
\Mg{(p,a),(p,b)}, \Mg{(p,b),(p,c)}, \Mg{(q,a),(q,b)} and  \Mg{(q,b),(q,c)}, 
OR $Bel_2$ must have focal points:
\Mg{(r,a),(r,b)}, \Mg{(r,b),(r,c)},
 \Mg{(s,a),(s,b),(t,a),(t,b)}  and     \Mg{(s,b),(s,c),(t,b),(t,c)}, 
OR BOTH. 

Let us suppose that 
$Bel_1$ has in fact focal points:
\Mg{(p,a),(p,b)}, \Mg{(p,b),(p,c)}, \Mg{(q,a),(q,b)} and  \Mg{(q,b),(q,c)}. 
Then $Bel_2$ must have neither 
\Mg{(r,a),(r,b)} nor \Mg{(r,b),(r,c)},
as focal point, because then \Mg{(p,r,b)} would be a focal point of $Bel$,
which is not the case. \\
But then $Bel_2$ has to have the focal point \Mg{(r,a),(r,b), (r,c)}.
Similarly, $Bel_2$ must have neither 
\Mg{(s,a),(s,b),(t,a),(t,b)} nor \Mg{(s,b),(s,c),(t,b),(t,c)},
as focal point, because then \Mg{(q,s,b), (q,t,b)} would be a focal point of
$Bel$, which is not the case. \\
But then $Bel_2$ has to have the focal point 
\Mg{(s,a),(s,b), (s,c), (t,a),(t,b), (t,c)}.
 Then, however, 
for the belief function $Bel$ we would have:
$$\frac{  m(\Mg{(p,r,a), (p,r,b)})}     
{         m(\Mg{(p,r,b), (p,r,c)})}
=
 \frac{  m(\Mg{ (p,s,a),(p,s,b),(p,t,a),(p,t,b)})}
{        m(\Mg{ (p,s,b),(p,s,c),(p,t,b),(p,t,c)})}
$$
which is not the case. In this way we arrive at a contradiction. 
We can reason by analogy reverting the roles of $Bel_1$ and 
$Bel_2$. Hence
it proves impossible to get the decomposition in terms of proper belief
functions.\\

\section{A Solution}

We define a new measure, beside m,Bel,Pl and Q, for the Dempster-Shafer
theory. Let r,s and v be three disjoint sets of variables. Let us restrict
our considerations to only those Bel functions, for which focal points 
are of the form: 
$$A=A ^{\downarrow r}\times A ^{\downarrow s}\times A ^{\downarrow v}$$.
One can call therefore each r,s and v by the term "dimension". 

What we are now interested in is the possibility of testing dependence
or independence of r and s, and later whether the dependence statement is
influenced by v. Given the relationship among r and s is influenced by v,
we may be interested, assuming causality among r,s,v, whether v makes r and s
independent.

(Unconditional) independence between r and s alone is trivial,
solvable with traditional statistical methods, as negative
mass values do not emerge in the process. The interesting case is that of
three variables.

Let us define the $F ^{r,s,v}$ function corresponding to a given belief
function Bel as 
\begin{equation}
 F ^{r,s,v}(A ^{\downarrow (r\cup s)}\times A ^{\downarrow v})=
\sum_{B; A ^{\downarrow v}\subseteq B}
m(A ^{\downarrow (r\cup s)}\times B)
\end{equation}

In an obvious way F measure differs significantly from the ordinary DST
mejasures in that it is a mixture of the Q-measure along the v dimension 
while the
m-measure along the r,s dimension. The function F is everywhere non-negative.

First of all we can test (main subject of the next section) whether v
influences relationship among r and s. 
If the relationship among variables sets r,s are not influenced by v, then if
the set R stems from space r, S from space s, and V from space v, then
$$F  ^{r,s,v}( R\times S\times V)=
m  ^{\downarrow r\cup s}( R\times S) \cdot F  ^{.,.,v}( V)$$

If the above equation is rejected, then conditional
independence of r,s given v may be of interest. 
At this point we need to assume existence of causal relationship 
of r,s on v.
If the variables sets r,s are independent given v, then if the set R stems
from space r, S from space s, and V from space v, then
$$F  ^{r,s,v}( R\times S\times V)=
F  ^{r,.,v}( R\times V) \cdot F  ^{.,s,v}( S\times V)$$
where the dot stands for the dimensions which is simply summed up 
(marginalized
in probabilistic sense). 

Appropriate direct statistical tests are not subject of this paper, 
but we can derive from the next section a stepwise procedure to 
check for conditional independence.
The above relationship suggests that we can test for independence given
variable
set v in that for every level of the variable set v we test independence of
variable sets r and s. Notice that in terms of frequencies at the given level
of v the objects (database records) counted in cells for different
combinations of levels of variables r and s are different, though same objects
may occur on different levels of variables v.

The concept of measure F allows for direct conditional independence testing
for DST using known statistical procedures. In the subsequent section the
details are described for the particular example of three variables: X (from
r), Y (from (s) and Z (from v).

\section{Database   Evaluation   of    Three-dimensional    Belief 
Distributions}

Assume that there are K non-zero values of $Bel ^{\downarrow  Z}$. 
It means that for K sets in the domain of variable Z corresponding 
mass m is non-zero.\\

Having database with records  which  are  representative,  in  the 
opinion of  the  researcher,  one  can  perform  the  traditional 
statistical analysis. It should be stressed, however,  that  this 
analysis depends on database not  only  for  practical,  empirical 
data. There exist some  aspects  of  statistical  analysis  which 
impose certain restrictions for simulated database either.


Now the problem is how  to  assess  the  structure  of  dependence 
between   Z   and   two-dimensional   belief   distribution   $Bel 
^{\downarrow \{X,Y\}}$.

The "realistic" sample sizes should be assured first. Sample  size 
in the number of records which corresponds to a  given  value  of 
$Bel ^{\downarrow Z}$. We suggest that  each  sample  size  should 
belong to the interval
\begin{equation} \label{Stern}
<6 \cdot N_{xy},25 \cdot N_{xy}>
\end{equation}
with $N_{xy}$ being the actual number of  cells  for  distribution 
$(X,Y)^\downarrow$.

$N_{xy}$ not always equals the product of $n_x\cdot n_y$ i.e. the 
product  of  the  possible  values  of  $X  ^\downarrow$  and   $Y 
^\downarrow$. Some so called structural zeros can exist.

If for a given value of $Z ^\downarrow$ the number of  records  in 
database does not belong to the interval  (\ref{Stern}),  then  we 
propose to recode variable Z.

A basic tool for a statistical analysis is a  hierarchical  model 
corresponding to the frequencies of  database  records.  We  will 
test the accuracy of the following expression:
\begin{equation}\label{SternStern}
ln E(F_{ijk})=f+\delta_i ^{(x)}+\delta_j ^{(y)} 
+\delta_k ^{(z)} + \delta_{ij}^{(xy)}
\end{equation}
where:\\
ln - natural logarithm,       \\
E - (statistical) expectation\\
$F_{ijk}$ - database frequency of records having i-th value  of  $X 
^\downarrow$, j-th value of $Y ^\downarrow$ and k-th value (or its 
superset) of $Z ^\downarrow$.\\
i=1,2,...,$n_x$, j=1,2,...,$n_y$, k=1,2,...,K.\\

The configuration of $\delta$-values of the right-hand side of the 
expression (\ref{SternStern}) have a meaningful interpretation.
Generally the belief variables X and Y can be mutually  dependent. 
The joint distribution of these two variables does not change,
however,  for 
different values of $Z ^\downarrow$ within  the  three-dimensional 
framework.

Parameters f and $\delta$ and the model as  a  whole  fulfill  the 
traditional  statistical  terminology.  $\delta$-values  are 
"contrasts" which means that all possible marginal sums of indexed 
$\delta$'s must equal zero.

Traditional way for checking the adequacy of (\ref{SternStern})  is 
through the $\chi ^2$  statistic.  $\delta$-values  are  estimated 
for this purpose. Appropriate number of "degrees of freedom"  must 
be taken into account  when  calculating  p-value  of  statistical 
significance.

Let us consider the  following  example.  $X  ^\downarrow$  and  $Y 
^\downarrow$ have only 2 possible values each. $Z ^\downarrow$ has 
4 values and there is no impossible combination of  3-dimensional 
discrete vectors, i.e. no structural zeros.

There are $4\cdot 2 \cdot 2$ degrees of freedom at the  beginning. 
We must subtract, however, 1 degree  on  behalf  of  the  constant 
 f and additionally 
$$1+1+3+1$$
degrees  for  subsequent  indexed  $\delta$-values,  taking   into 
account the marginal restrictions.

The $\chi ^2$ statistic has in our example 9 degrees  of  freedom. 
If there would be some impossible triples of $(X,Y,Z)$, then  9  is 
diminished still: one degree for each of them.

Taking into account the "realistic" sample sizes  we  can  find  a 
first assessment of the possibility to factorize 
$Bel(X,Y,Z)$ with respect to its last component: $Z$. 

We propose the general threshold for the  p-values  calculated  for 
$\chi ^2$ statistic, to be p=0.1.

If the actual p-value is  smaller  than  0.1,  one  can  still  be 
seeking a factorization by redefinition of  the  variable  Z.  The 
notion of standardized residual can be used for this purpose.

The $\chi ^2$ statistic is the sum of squares of residuals of  the 
form:
$$\frac{F_{ijk}-\hat{F}_{ijk}}{\sqrt{\hat{F}_{ijk}}}$$
where $\hat{F}_{ijk}$ is fitted frequency.

To fit frequencies  according  to  model  (\ref{SternStern}),  the 
standard statistical programs can  be  employed  (e.g.  Statistica 
\cite{Statistica}, SPSS \cite{SPSS}).

Clearly those values of $Z ^\downarrow$ with the highest residuals 
can be starting points to redefinition of  this  variable.  Taking 
into account matrix  differences  in  estimated  probabilities  of 
$(X,Y)$  for  given  value  of  $Z  ^\downarrow$  one  can  obtain 
classification being a basis of a set of  redefined  variables  $Z 
^\downarrow$.

New $(X,Y,Z_1), (X,Y,Z_2),...$ fulfill  appropriateness  of  model 
(\ref{SternStern}), i.e. p-values calculated for restricted  $\chi 
^2$ statistics are higher than 0.1.

Marginals $Z_1^\downarrow$, $Z_2^\downarrow$, $Z_3^\downarrow$, ... can have
disjoint sets of values or not. For the real data the criterion of
meaningfulness must be taken into account when choosing among possible
triples.

Problem of joining of sets of probabilities is considered  in 
literature. Most recent publication  of  this  kind  is  Consonni, 
Veronese \cite{Consonni}. 

We have assumed that it is still easier to prove the existence  of 
differences in probabilities if certain elements of samples are in 
common.
Once  the  distribution  (X,Y,Z')   has   passed   the   test   for 
factorization, one must confirm it. 

The  problem  of  comparing  the  contingency  tables  which  were 
generated by the samples having some elements in  common  is  less 
addressed in the statistical literature. Only recently
\cite{Thomson:95} the  optimal  variance  for  difference  of  two 
proportions was calculated.

Previously the same problem was considered from different point of 
view
\cite{Choi,Shih}. The missing data  was  allowed  in  the  matched 
(paired) experiment for proportions.

Additional aspects of the procedure just described can be found in
\cite{Agresti:92,Agresti:95,Silva:95}.

\end{document}